\documentclass{article}
\usepackage{ijcai21}

\usepackage{times}

\usepackage{soul}
\usepackage{url}
\usepackage[hidelinks]{hyperref}
\usepackage[utf8]{inputenc}
\usepackage[small]{caption}
\usepackage{graphicx}
\usepackage{amsmath}
\usepackage{booktabs}
\usepackage{color}

\usepackage{multirow}
\usepackage{multicol}
\usepackage{graphicx}
\usepackage{subfigure}
\usepackage{makecell}

\usepackage{amssymb}
\usepackage{caption}
\usepackage{graphicx,subfigure}
\usepackage{booktabs,amsfonts,dcolumn}

\newcommand{\blue}[1]{\textcolor{blue}{#1}}

\def\mathbi#1{\textbf{\em #1}}

\title{KDExplainer: A Task-oriented Attention Model for Explaining Knowledge Distillation}

\author{
Mengqi Xue$^1$\and
Jie Song$^1$\and
Xinchao Wang$^{2}$\and
Ying Chen$^1$ \and
Xingen Wang$^1$\footnote{Corresponding author} \and
Mingli Song$^1$
\\
\affiliations
$^1$Zhejiang University,
$^2$National University of Singapore\\
\emails
\{mqxue, sjie, lynesychen, newroot, brooksong\}@zju.edu.cn,
xinchao@nus.edu.sg

}

\begin{document}

\maketitle

\begin{abstract}
Knowledge distillation~(KD) has recently emerged
as an efficacious scheme for
learning compact deep neural networks~(DNNs).
Despite the promising results achieved,
the rationale that interprets
the behavior of KD has yet remained largely
understudied. In this paper,
we introduce a novel task-oriented attention model, 
termed as KDExplainer, to shed light on the working mechanism
underlying the vanilla KD.
At the heart of KDExplainer is a 
Hierarchical Mixture of Experts (HME),
in which a multi-class classification 
is reformulated as a multi-task binary one.
Through distilling knowledge from a free-form
pre-trained DNN to KDExplainer,
we observe that KD 
implicitly modulates the knowledge 
conflicts between different subtasks,
and in reality has much more to offer
than label smoothing. 
Based on such findings,
we further introduce a portable
tool, dubbed as virtual attention module~(VAM),
that can be seamlessly integrated with 
various DNNs to enhance their performance under KD.  
Experimental results demonstrate that
with a negligible additional cost, 
student models equipped with VAM consistently 
outperform their non-VAM counterparts across different benchmarks.
Furthermore, when combined with other KD methods, 
VAM remains competent in 
promoting results,
even though it is only motivated by 
vanilla KD. The  code  is  available at\textit{~\url{https://github.com/zju-vipa/KDExplainer}}.

\end{abstract}

\section{Introduction}

Recent progress in deep neural networks (DNNs) 
has significantly benefited 
many if not all tasks in artificial intelligence,
ranging from computer vision to
natural language processing. 
The encouraging results, however,
come with an enormous cost:
state-of-the-art DNNs have been scaled up to hundreds and even thousands of
layers with millions of parameters, 
making it demanding to
train and deploy even on GPU clusters, let alone 
on edge devices like mobile phones. 

Many research efforts have thus
been made to craft lightweight DNNs 
applicable to in-the-wild scenarios.
Representative schemes include  
weight pruning~\cite{Han2015LearningBW}, 
model quantization~\cite{Jacob2018QuantizationAT}, 
and knowledge distillation~(KD)~\cite{Hinton2015DistillingTK},
among which KD has recently emerged as one
of the most flourishing topics in
the field. 
The goal of KD is to extract 
knowledge from a well-behaved but
cumbersome teacher model,
often known as \emph{dark knowledge},
and to learn a compact student model 
capable to handle the task of the teacher
but with fewer parameters.
Since the pioneering work of \cite{Hinton2015DistillingTK}, 
a variety of dark knowledge has been explored, 
including \emph{hint}~\cite{Romero2015FitNetsHF}, 
\emph{attention}~\cite{Zagoruyko2017AT}, and 
\emph{instance relationship}~\cite{Liu2019KnowledgeDV}. 

Despite the above forms of dark knowledge showcase promising results, 
the naive \emph{soft target}~\cite{Hinton2015DistillingTK} 
is found still among the most competitive ones~\cite{tian2019crd}. 
Nevertheless, few attempts have been dedicated to 
explaining the rationale of soft targets on the student learning.  
A common belief is that soft labels reveal richer 
information like category similarities than the widely-used one-hot vector, 
so that the student obtains more supervision signals for learning. 
Recently, the work~\cite{Yuan_2020_CVPR} argues that the soft target is intrinsically a type of 
learned label smoothing~\cite{Szegedy2016RethinkingTI} 
regularization. ~\cite{Furlanello2018BornAN} conjectures 
that soft targets resemble importance-weighting, 
where weights correspond to confidences
of the teachers in the correct prediction. 
~\cite{Tang2020UnderstandingAI},
on the other hand, 
dissects the effects of soft targets into three main factors: 
label smoothing, importance weighting, and category similarities. 
Albeit the inspiring insights provided by these works, 
the underlying mechanism of how category similarities 
regularize the student model learning 
has remained largely under-studied to date.


In this paper, we take a closer look at the role of the soft label 
through the lens of a novel attention model,  which we term as KDExplainer. KDExplainer is an interpretation-friendly 
student model, which distinguishes itself from conventional student DNNs
from two main aspects, as illustrated in  Figure~\ref{fig:overview}.
First,  KDExplainer takes the form of 
a Hierarchical Mixture of Experts (HME), where each expert
is expected to specialize in specific subtasks and
learn adaptive features. 
Second, KDExplainer casts the multi-class classification task as a multi-task problem, 
in which each task is formulated as a binary classification problem.
Such a design explicitly 
shapes KDExplainer as a 
neural tree,
for which the inherent working mechanism
is well understood,
in the aim to interpret KD.

We then carry out KD from a free-form pre-trained DNN to the dedicated
KDExplainer, through which process the rationale
of KD is highlighted thanks to the 
interpretable essence of the HME architecture. 
Interestingly, we find that the KD objective
promotes lower entropy of the attention distribution, 
indicating that soft labels, in reality,
encourage specialization of different experts 
and hence play a role in modulating the knowledge
conflicts for solving different subtasks. 
To further understand  
the connection and difference between KD and 
label smoothing, 
we train a KDExplainer using label smoothing,
and discover that the derived 
attention distribution 
exhibits no significant differences with those by vanilla training without KD.
This phenomenon marks that soft labels indeed have more to offer,
including feature specialization,
than label smoothing.

Inspired by these observations, 
we further introduce a portable component, 
termed as virtual attention module (VAM),
to enhance the performance of conventional DNNs under KD. 
The key idea of VAM is to coordinate
the knowledge conflicts for discriminating different categories,
achieved via lowering the entropy of the attention distribution.
VAM can be readily integrated with existing DNNs
while bringing negligible additional computation cost.
Moreover, since VAM is naturally orthogonal to KD, 
it can be seamlessly combined with various KD schemes,
rendering it a handy module
to facilitate KD.

Our contributions are therefore summarized as follows.
\begin{itemize}
    \item We propose a novel attention model of 
    interpretable nature, KDExplainer, 
    to understand the role of soft labels 
    in training the student.
    \item Through KDExplainer, we observe that soft labels 
    implicitly modulate the knowledge 
    conflicts between different subtasks by promoting feature specialization,
    and offer more regularization
    than only label smoothing.
    \item Understanding the KD rationale via KDExplainer further motivates us to design  a portable and compact module,
    VAM, readily applicable to various DNNs and KDs.
\end{itemize}   
Extensive experimental results across benchmarks 
demonstrate that VAM not only consistently improves
the performance of vanilla KD under various experimental settings,
but also can be readily integrated with 
other state-of-the-art KD schemes to further promote their results.

\section{Related Work}
\paragraph{Knowledge Distillation}
Knowledge distillation has attracted increasing attention thanks to its important role in deploying deep networks to low-capacity edge devices. The main idea is leveraging the \textit{dark knowledge} encoded in a bulky teacher to craft a lightweight student model with performance on par with the teacher. Over the last several years, most works devote themselves to the exploration of different forms of the dark knowledge, including soft targets~\cite{Hinton2015DistillingTK}, features~\cite{Romero2015FitNetsHF}, attention~\cite{Zagoruyko2017AT}, factors~\cite{NIPS2018_7541}, activation boundary~\cite{ABdistill}, and instance relationship~\cite{Liu2019KnowledgeDV,Park2019RelationalKD,tung2019similarity}. By imitating the teacher to behave in a similar way, the student achieves comparable performance even with much fewer parameters.

\paragraph{Attention Mechanism}
Inspired by human cognition, attention mechanisms  focus on relevant regions of input data to solve the desired task rather than ingesting the entire input. Attention-based neural networks have been broadly adopted in natural language models for machine translation~\cite{Bahdanau2015NeuralMT}, image caption generation~\cite{Xu2015ShowAA}, and unsupervised representation learning~\cite{Devlin2019BERTPO}. Attention mechanisms also achieve great success in vision models~\cite{Mnih2014RecurrentMO,Bello_2019_ICCV}. Except the performance boost, attention mechanism also provides an important way to explain the workings of neural models~\cite{Li2016UnderstandingNN,Wiegreffe2019AttentionIN,Xu2015ShowAA}. 
Unlike most prior works, 
our focus here is to utilize an 
attention mechanism to interpret KD, which has been largely overlooked in previous literature. 

\section{KDExplainer}

\begin{figure*}
     \centering
    \includegraphics[scale=0.65]{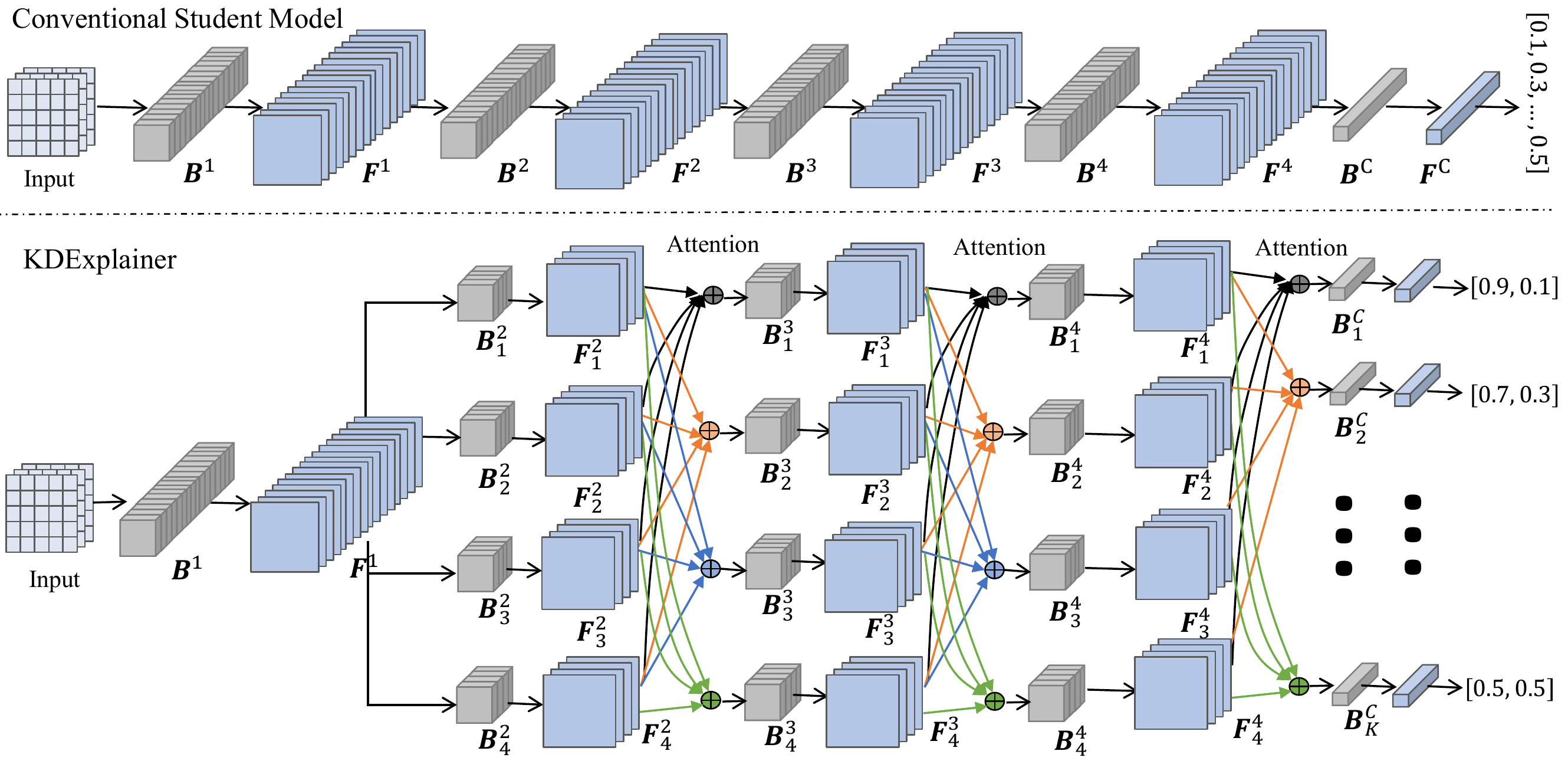}
    \vspace{-0.5em}
    \caption{A conceptual illustration. \textbf{Top}: a conventional student network for knowledge distillation. \textbf{Bottom}: the proposed KDExplainer. To explain the effects of soft targets as the distillation objective, we use the KDExplainer as the student model for knowledge distillation. }
    \label{fig:overview}
    \vspace{-1em}
\end{figure*}

\subsection{Knowledge Distillation with Soft Targets}
Vanilla KD~\cite{Hinton2015DistillingTK} distills the ``dark knowledge'' from the teacher via aligning the soft targets \begin{equation}\label{eq:vanilla_distill}
\mathcal{O}_{\text{KD}} =\alpha \mathcal{L}_{\text{CE}}(p(\mathbi{z}^s), \mathbi{y}) + (1-\alpha) \mathcal{D}_{\text{KL}}\left(p(\mathbi{z}^t; {\tau}), p(\mathbi{z}^s; {\tau})\right),
\end{equation}
where $\mathbi{z}^s$ and $\mathbi{z}^t$ are respectively the logits from the student and the teacher models.  $\mathbi{y}$ is its associated one-hot label vector, $p$ denotes the softmax function that produces the category probabilities given the logits, and
$\tau$ is a non-negative temperature hyperparameter  used to smooth the distributions. As for $p_i(\mathbi{z};\tau)$, we have
\begin{equation}\label{eq:soft_target}
  p_i(\mathbi{z};\tau)=\frac{\exp({z}_i/\tau)}
  {\sum\nolimits_{j}\exp({z}_j/\tau)}.
\end{equation}
Also, $\mathcal{L}_{\text{CE}}$ is the conventional cross-entropy loss, and $\mathcal{D}_{\text{KL}}$ is the Kullback-Leibler divergence between the categorial distributions predicted from the teacher and the student models. $\alpha$ is a hyperparameter to trade off the two objective terms.

\subsection{The Proposed KDExplainer}

We propose the KDExplainer, a task-oriented attention model, as the student model to uncover the rationale
underlying KD. 
KDExplainer makes two main modifications based on exiting popular DNNs: (1) dissecting the effects of class similarity in soft targets, KDExplainer reformulates the multi-class classification problem as an ensemble of multiple binary classification problems; (2) KDExplainer remodels the student model as a Hierarchical Mixture of Experts (HME) and introduces a task-oriented attention mechanism as the gating function to make the student model more friendly for human interpretation. The overview of the proposed KDExplainer is shown in Figure~\ref{fig:overview}. KDExplainer is designed based on existing widely-used networks such as ResNet~\cite{he2016deep}, Wide Residual Network~\cite{zagoruyko2016wide}, and VGG~\cite{simonyan2014very}. These classic DNNs have some common characteristics: these models are usually composed by several blocks, each of which is a stack of convolutional layers, batch normalization layers~\cite{ioffe2015batch}, and nonlinear activation layers. The number of filters usually keeps fixed in the same block and changes across different blocks. 

Formally, we use $B^i$ to denote the $i$-th block, and
then a standard DNN can be formulated as $F_{DNN} = B^C\circ B^{L}\circ\cdot\cdot\cdot \circ B^2\circ B^1$, where symbol $\circ$ denotes the function composition operation. $B^C$ is the classification block that consists of the fully connected layer and the softmax layer. KDExplainer roughly follows the design of the existing DNN, but divides each block $B^i$ (except the first block $B^1$) into $N_i$ equal-sized sub-blocks, \textit{i.e.}, $\hat{B}^i=\{B^i_1, B^i_2, ..., B^i_{N_i}\}$ for any $i>1$. We view each sub-block $B^i_j$ as an expert, and introduce a task-oriented attention module before each expert as the gating function to select a combination of the outputs from previous experts. Therefore, the whole model can be viewed as an HME model.

\subsubsection{Task-oriented Attention}
The proposed attention is ``task-oriented'' as the parameters of the attention module are trained on the whole dataset, which is largely different from the attention mechanism in existing literature~\cite{Bahdanau2015NeuralMT,Bello_2019_ICCV} where the attention weights are determined by instances. For each sub-block $B^i_j$, we use a trainable parameter vector $\mathbi{v}^i_j=[v^i_{j,1}, v^i_{j,2}, ..., v^i_{j,N_{i-1}}]$ to learn the attention distribution over previous experts. 
Let output feature maps from previous blocks be $\left\{\mathbi{F}^{i-1}_1,...,\mathbi{F}^{i-1}_{N_{i-1}}\right\}$. At training phase, the input of sub-block $B^i_j$ is computed by
\begin{equation}
    \widetilde{\mathbi{F}^i_{j}} = \sum\nolimits^{N_{i-1}}_{k=1}a^i_{j,k}\cdot\mathbi{F}^{i-1}_{k},
\end{equation}
where $a^i_{j,k}$ is the attention weight of the $k$-th expert, $a^i_{j,k} = \frac{\exp({v^i_{j,k}}/T)}{\sum_m \exp({v^i_{j,m}/T})}$. $T$ is a temperature hyper-parameter shared by all attention modules. Note that the classification block $B^C$ is divided into $K$ blocks, where $K$ equals the number of classes in the classification problem. The multi-class classification problem thus turns into an ensemble of binary classification problems in KDExplainer, as shown in Figure~\ref{fig:overview}. 

\subsubsection{Explaining KD by KDExplainer}
Recall that KDExplainer is a substitute of the conventional student model, using KDExplainer to understand the effects of KD is straightforward: analyzing differences between the experimental results of KDExplainer trained with and without KD.
As KDExplainer reformulates the multi-class classification as multiple binary classification tasks, each binary classification task encounters the imbalanced data problem. Let $\mathbi{p}^s_k=[p^s_{k,0}, p^s_{k, 1}]$ be the probability predictions of KDExplainer for the $k$-th classification task, and $\mathbi{y}_k=[y_{k,0}, y_{k, 1}]$ be the ground truth in the form of a one-hot vector. When the proposed KDExplainer is trained with KD, the objective function is
\begin{equation}
\label{eq:KDEobj}
\begin{aligned}
\mathcal{O}_{\text{KD}} =\sum\nolimits_{k=1}^K \Big \{ 
\alpha \mathcal{L}_{\text{WCE}}\left(\mathbi{p}^s_k, \mathbi{y}_k\right) + 
(1-\alpha)  \mathcal{D}_{\text{KL}}\left(\hat{\mathbi{q}}^t_k, \hat{\mathbi{p}}^s_k\right)
\Big \},
\end{aligned}
\end{equation}
where $\mathcal{L}_{\text{WCE}}$ is the weighted cross-entropy loss 
\begin{equation}
\mathcal{L}_{\text{WCE}}=-w_0 y_{k,0}\log p^s_{k,0}-w_1y_{k,1}\log p^s_{k,1}.
\end{equation}
$w_0$ and $w_1$ are the weights balancing the cost incurred by the positive and the negative samples, which alleviates the negative effects caused by imbalanced data. $\hat{\mathbi{q}}^t_k$ and $\hat{\mathbi{p}}^s_k$ are the softened category probabilities from the teacher and the student models, respectively. Note that all teacher models involved in this paper are still conventional DNNs for multi-class classification, so we convert the $K$-class probability prediction $\mathbi{p}^t=[p^t_0,p^t_1, ..., p^t_K]$ from the teacher to $K$ two-dimensional probability vectors as follows
\begin{equation}
\hat{\mathbi{q}}^t_k = [1-p^t_k, p^t_k], \  \text{for any}\ k \in \{1, 2, ..., K\}.
\end{equation}
If KDEplainer is trained without KD, the second term in Eqn.~\ref{eq:KDEobj} is removed. 

KDExplainer enjoys an appealing property thanks to the elaborate design: after training, it becomes a tree-like multi-branch network (i.e., neural tree) if only the maximum in the attention weights is kept:
\begin{equation}
\label{eq:keep-maximum}
   a^i_{j,k}= \left\{
   \begin{aligned}
        &1, \ \text{if}\ k=\arg\max_m a^i_{j, m}; \\
        &0, \    \text{otherwise}.
    \end{aligned}
   \right.
\end{equation}
The derived neural trees provide us with stronger support for interpretation and analysis of knowledge distillation.

\section{Virtual Attention Mechanism}
\begin{figure}[t]
     \centering
    \includegraphics[scale=0.51]{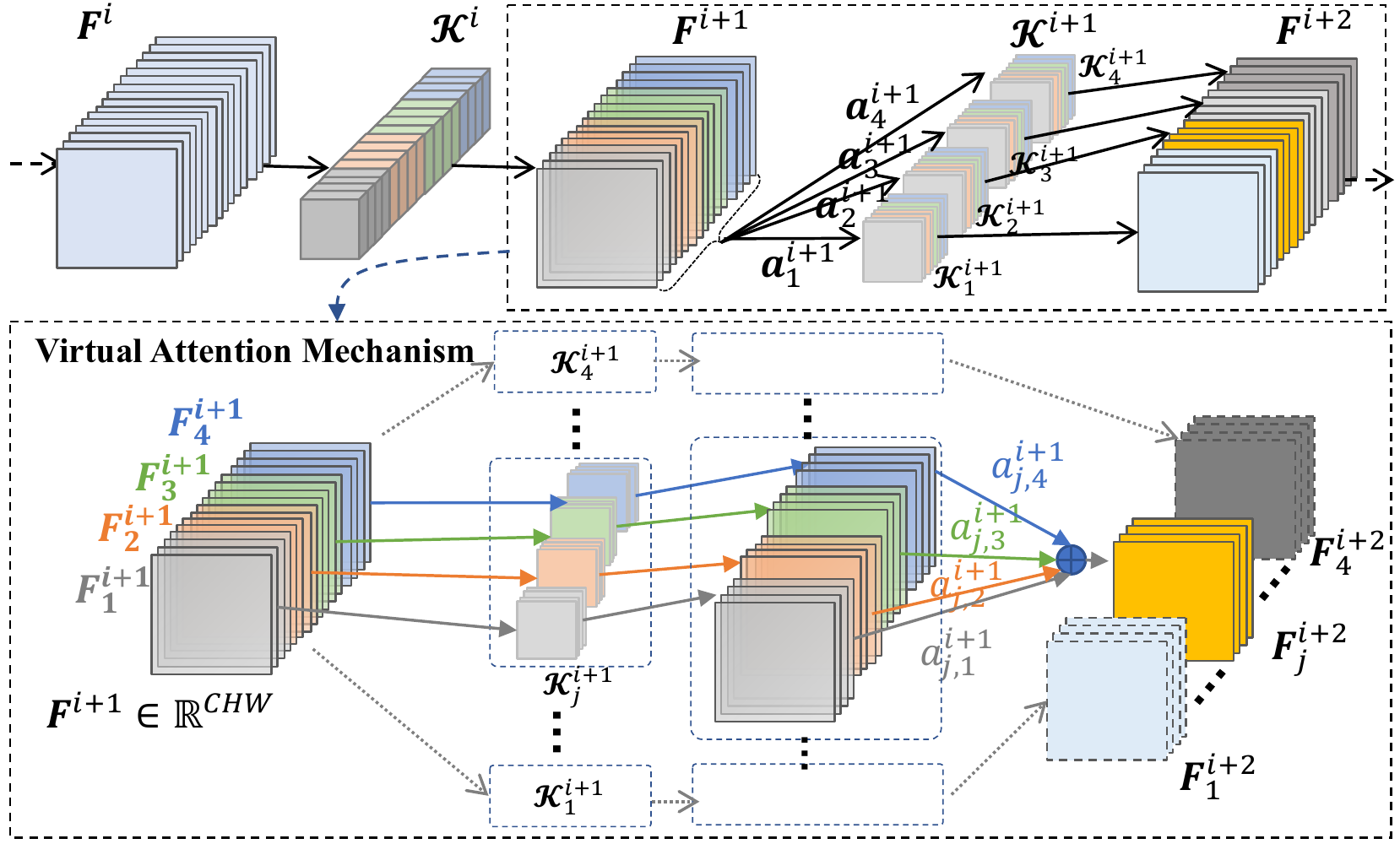}
    \caption{The proposed virtual attention mechanism. For simplicity, here we only depict the details of how $\mathbi{F}^{i+1}$ is convolved with the virtual filter block $\mathcal{K}^{i+1}_j$ to produce $\mathbi{F}^{i+2}_j$.}
    \label{fig:vam}
    \vspace{-1em}
\end{figure}
KDExplainer is tailored to understand the effects of soft targets in KD. However, it may suffer from slight accuracy sacrifice compared to conventional student DNNs due to the \textit{ad hoc} design that multi-class classification turns into multiple binary classification tasks. Here we propose a virtual attention mechanism that is easily integrated into existing student models with few architecture modifications, to retain the capacity of these student models for higher KD performance. As shown in Figure 2, VAM views all convolution filters in a layer as several ``virtual'' filter blocks, each filter block akin to the expert in the KDExplainer. Note that ``virtual'' means that VAM does not really split the convolution filters into any blocks. The only modification is that VAM slightly changes the conventional convolution operation by incorporating an attention mechanism.

Formally, we use $\mathcal{K}^i\in \mathbb{R}^{C^i_{out}C^i_{in}S S}$ to denote the convolution filters in the $i$-th layer, where 
$C^i_{in}$ and $C^i_{out}$ denote the input and the output channels. Here for simplicity, we assume all filters are square, where both the width and the length are $S$. Assume the input tensor of the $i$-th convolution layer is ${\mathbi{F}^i}\in \mathbb{R}^{C^i_{in}HW}$, where $H$ and $W$ are the height and the width of the feature map. After the $i$-th layer and the $(i+1)$-th layer, the features turn into $\mathbi{F}^{i+1}\in \mathbb{R}^{C^i_{out}HW}$ and $\mathbi{F}^{i+2}\in \mathbb{R}^{C^{i+1}_{out}HW}$, respectively.
\begin{figure*}[ht]
     \centering
     \subfigure[Normal]{\includegraphics[scale=0.38]{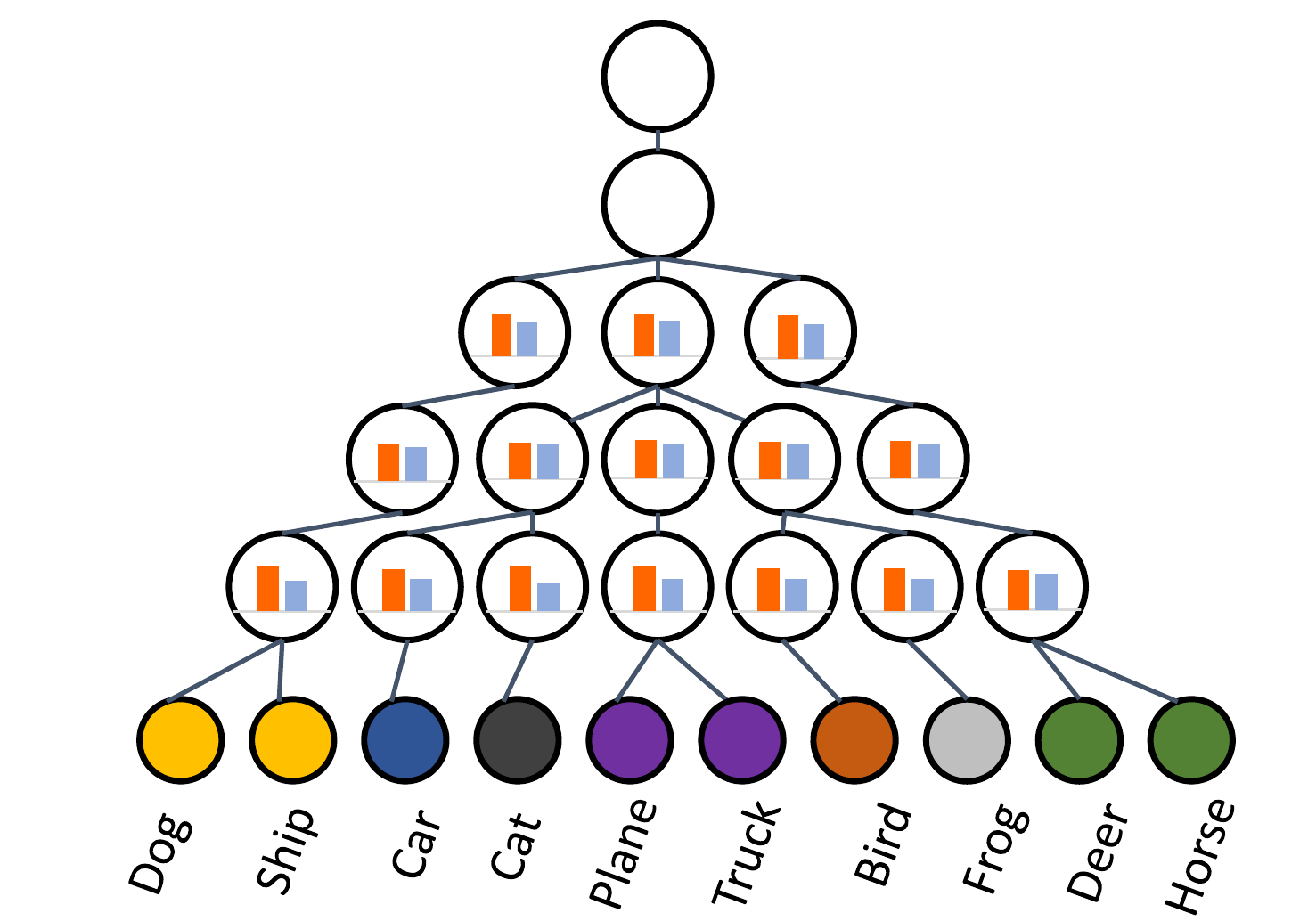}}
     \subfigure[Label Smoothing]{\includegraphics[scale=0.38]{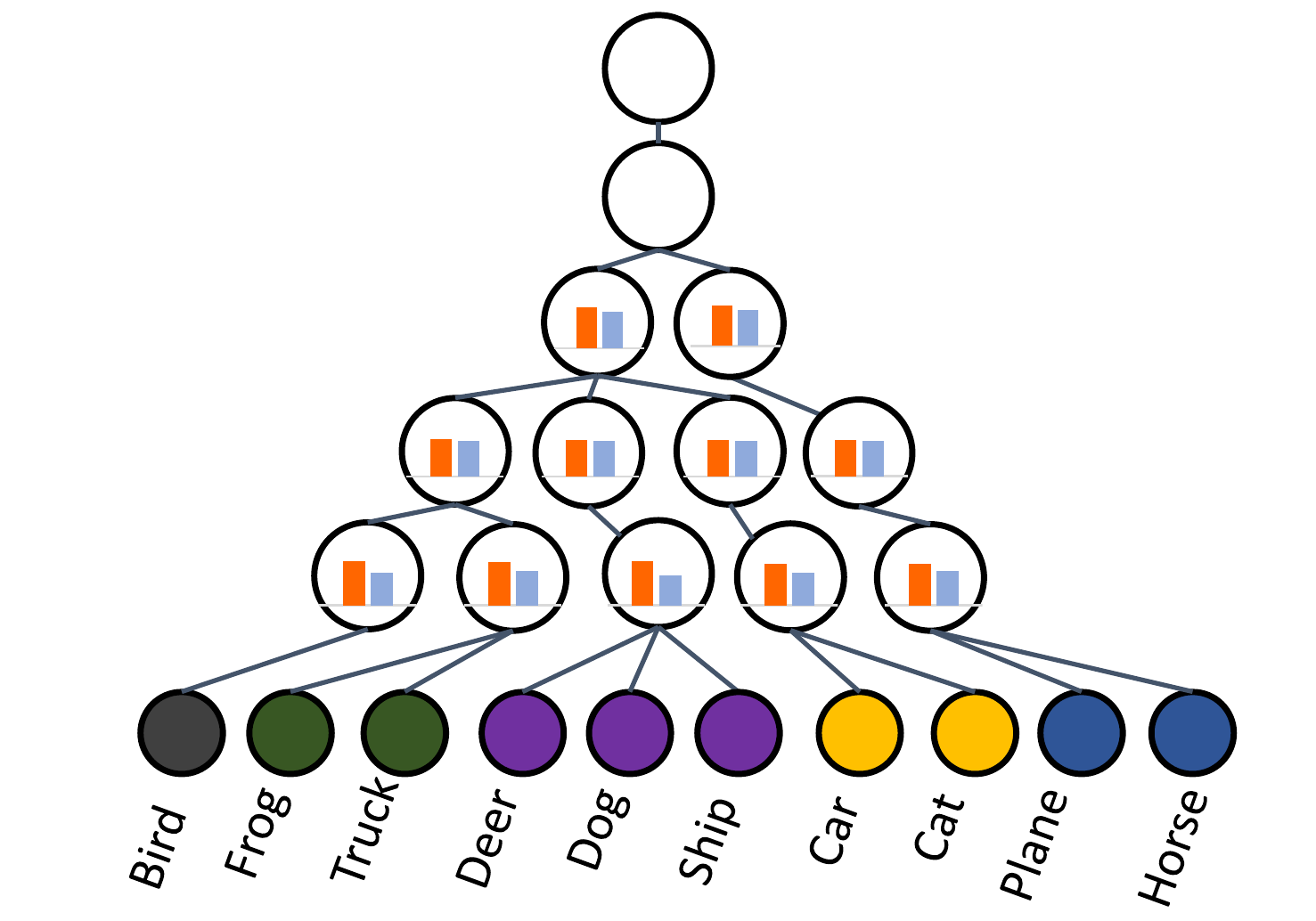}}
     \subfigure[KD]{\includegraphics[scale=0.38]{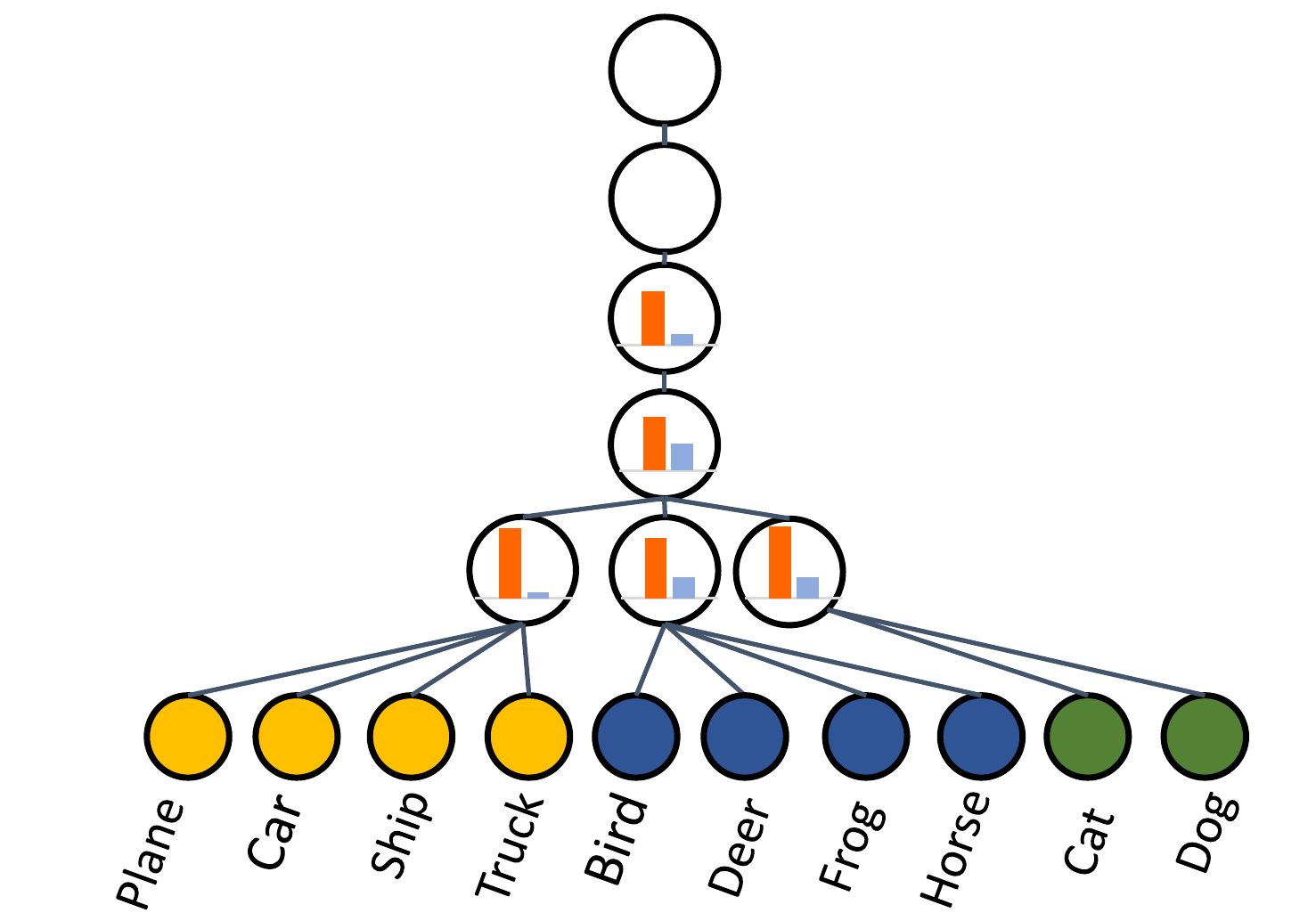}}
    \vspace{-1em}
    \caption{Visualization of the derived neural trees from the proposed KDExplainer, which is designed based on ResNet18 and trained on CIFAR-10. (a) KDExplainer trained with normal cross-entropy loss; (b) KDExplainer trained with label smoothing; (c) KDExplainer trained with KD. Each node denotes the retained block by removing the unused blocks according to Eqn~\ref{eq:keep-maximum}. The histogram in each node represents the attention distribution over its input tensors. The red and the blue bars denote the maximum and the minimum, respectively.}
    \label{fig:experiment-tree}
    \vspace{-1em}
\end{figure*}

To incorporate the attention module into the DNN, we view convolution filters in each layer as a set of filter blocks. Assume the filters in the $i$-th layer are divided into $M$ virtual groups, such that $\mathcal{K}^{i}=\{\mathcal{K}^{i}_1, \mathcal{K}^{i}_2, ..., \mathcal{K}^{i}_M\}, \mathcal{K}^{i}_j\in \mathbb{R}^{(C^i_{out}/M)C^i_{in}S S}$ for $1\le j\le M$. After the $i$-th layer, the feature maps produced by each block can be calculated as follows
\begin{equation}
   \mathbi{F}^{i+1}_j = \mathbi{F}^{i}\odot \mathcal{K}^i_j,  \  \text{for any}\ j \in \{1, 2, ..., M\},
\end{equation}
where $\odot$ denotes the convolution operation. The whole feature tensor can be denoted by 
\begin{equation}
   \mathbi{F}^{i+1} = \coprod\nolimits^{M}_{m=1}\mathbi{F}^{i+1}_m, 
\end{equation}
where $\coprod$ denotes feature concatenation.  Assume filters in the $(i+1)$-th layer are divided into $N$ virtual blocks, such that $\mathcal{K}^{i+1}=\{\mathcal{K}^{i+1}_{1},\mathcal{K}^{i+1}_{2},..., \mathcal{K}^{i+1}_{N}\}, \mathcal{K}^{i+1}_{j}\in\mathbb{R}^{(C^{i+1}_{out}/N)C^{i+1}_{in}S S}$ for $1\le j\le N$. To align each filter block in the $(i+1)$-th layer with the output from each block of the $i$-th layer, we further divide each block $\mathcal{K}^{i+1}_{j}$ into $M$ groups along the input channel, such that $\mathcal{K}^{i+1}_{j}=\{\mathcal{K}^{i+1}_{j,1}, \mathcal{K}^{i+1}_{j,2}, ..., \mathcal{K}^{i+1}_{j, M}\}, \mathcal{K}^{i+1}_{j,l}\in\mathbb{R}^{(C^{i+1}_{out}/N)(C^{i+1}_{in}/M)S S}$ for any $1\le l\le M$. Similar to KDExplainer, we introduce a task-oriented attention module $\mathbi{a}^{i+1}_j=\left[a^{i+1}_{j,1}, a^{i+1}_{j,2}, ..., a^{i+1}_{j,M}\right]$ before each filter block $\mathcal{K}^{i+1}_{j}$, thus the output of the block can be computed by 
\begin{equation}
   \mathbi{F}^{i+2}_j = \sum\nolimits^{M}_{m=1}a^{i+1}_{j,m}\mathbi{F}^{i+1}_{m}\odot\mathcal{K}^{i+1}_{j, m}.
\end{equation}
The attention weights are computed by a softmax function over trainable parameters $\mathbi{v}=\left[v^{i+1}_{j,1},..., v^{i+1}_{j,M}\right]$, i.e., 
\begin{equation}
    a^{i+1}_{j,m} = \frac{\exp{(v^{i+1}_{j,m})}}{\sum\nolimits^{M}_{k=1}\exp{}(v^{i+1}_{j,k})}.
\end{equation}
As soft targets are found to encourage lower entropy of the attention distribution (seen in Section~\ref{subsec:experiment-KDE}), we introduce another regularization term based on the conventional KD objective in Eqn.~\ref{eq:vanilla_distill}
\begin{equation}\label{eq:vam_obj}
\begin{aligned}
\mathcal{O}_{\text{KD}} = & (1-\alpha) \mathcal{D}_{\text{KL}}\left(p(\mathbi{z}^t; {\tau}), p(\mathbi{z}^s; {\tau})\right) + \\
& \alpha \mathcal{L}_{\text{CE}}(p(\mathbi{z}^s), \mathbi{y}) + \gamma\mathcal{H}(A),
\end{aligned}
\end{equation}
where $A$ denotes all the involved attention distributions, and $\mathcal{H}$ is the sum of their entropy
\begin{equation}
\label{eq:entropy-minimize}
    \mathcal{H}(A)=\sum\nolimits_{i}\sum\nolimits_{j}\sum\nolimits_k-a^{i}_{j,k}\log{a^i_j,k}.
\end{equation}

\section{Experiments}

\subsection{Explaining KD with KDExplainer}
\label{subsec:experiment-KDE}

\subsubsection{Experimental settings} Experiments are conducted on CIFAR-10 and CIFAR-100~\cite{krizhevsky2009learning}. We implement the proposed 
KDExplainers based on four DNN architectures: ResNet18~\cite{he2016deep}, VGG8~\cite{simonyan2014very}, WRN-16-2, and WRN-40-1~\cite{zagoruyko2016wide}. For all models, we use ResNet50 as their teacher model for KD. The initial learning rate is 0.1 and decayed every 30 epochs. The training ceases at 300 epochs. For more details, please refer to supplementary materials.

\subsubsection{Experimental results}
\begin{table*}[t]
    \centering
    \resizebox{\textwidth}{!}{
    \begin{tabular}{c|cccccc|cccccc}
      \toprule
      \textbf{Data} &\textbf{Model}& \textbf{P-DNN} &\textbf{VAM} &$\mathcal{L}_{\text{CE}}$ & $\mathcal{D}_{\text{KL}}$ &$\mathcal{H}$ & \textbf{VGG8} & \textbf{WRN-16-2} & \textbf{WRN-40-1} & \textbf{resnet20} & \textbf{ResNet18} & \textbf{ShufNetV1} \\  \midrule
      \multirow{6}{*}{\rotatebox{90}{CIFAR-10}} &M1&$\checkmark$
      &&$\checkmark$&& & 91.41 & 93.71 & 93.34 & 92.83 & 95.26 & 92.56   \\
      &M2&&$\checkmark$&$\checkmark$&& & 91.69 & 93.46 & 93.47 & 92.56 & \blue{95.51} & 92.82  \\
      &M3&&$\checkmark$&$\checkmark$&& $\checkmark$& 91.82 & 93.90 & 93.95 & 92.86 & 94.99 & 92.83  \\
      &M4&$\checkmark$&&$\checkmark$&$\checkmark$& & 93.14 & 94.55 & 93.86 & 93.08 & 95.43 & 93.50 \\
      &M5&&$\checkmark$&$\checkmark$&$\checkmark$& & \blue{93.19} & \blue{94.67} & \blue{93.96} & \blue{93.39} & \blue{95.51} & \blue{93.70} \\
      &M6&&$\checkmark$&$\checkmark$&$\checkmark$&$\checkmark$ & \textbf{93.36}  & \textbf{94.85} & \textbf{94.32} & \textbf{93.50} & \textbf{95.59} & \textbf{93.79} \\

      \midrule
      \multirow{6}{*}{\rotatebox{90}{CIFAR-100}} &M1&$\checkmark$
      &&$\checkmark$&& & 70.37 & 73.15 & 71.36 & 69.84 & 77.18 & 71.45  \\
      &M2&&$\checkmark$&$\checkmark$&& & 70.54 & 73.56 & 71.15 & 69.58 & 76.62 & 71.70 \\
      &M3&&$\checkmark$&$\checkmark$&& $\checkmark$& 70.98 & 73.92 & 71.61 & 69.71 & 78.30 & 71.81 \\
      &M4&$\checkmark$&&$\checkmark$&$\checkmark$& & 73.53 & 75.01 & 73.44 & 70.05 & 79.54 & 75.41 \\
      &M5&&$\checkmark$&$\checkmark$&$\checkmark$& & \blue{73.78} & \blue{75.43} & \blue{73.87} & \blue{70.32} & \blue{79.63} & \blue{75.62} \\
      &M6&&$\checkmark$&$\checkmark$&$\checkmark$&$\checkmark$ & \textbf{74.17} & \textbf{75.63} & \textbf{73.92} & \textbf{70.43} & \textbf{79.77} & \textbf{76.10} \\

      \midrule
      \multirow{6}{*}{\rotatebox{90}{Tiny-ImageNet}} &M1&$\checkmark$
      &&$\checkmark$&& & 56.47 & 57.52 & 56.26 & 50.54 & 65.59 & 60.52 \\
      &M2&&$\checkmark$&$\checkmark$&& & 57.18 & \blue{58.16} & 56.02 & 50.53 & 65.79 & 61.24 \\
      &M3&&$\checkmark$&$\checkmark$&& $\checkmark$& 57.63 & 58.08 & 56.23 & 50.70 & 66.33 & 62.47 \\
      &M4&$\checkmark$&&$\checkmark$&$\checkmark$& & 60.41 & 57.91 & 56.39 & 52.78 & 69.37 & \blue{65.54} \\
      &M5&&$\checkmark$&$\checkmark$&$\checkmark$& & \blue{60.45} & \textbf{58.24} & \blue{56.80} & \blue{52.83} & \blue{69.63} & 65.32 \\
      &M6&&$\checkmark$&$\checkmark$&$\checkmark$& $\checkmark$& \textbf{60.57} & 58.01 & \textbf{56.99} & \textbf{53.54} & \textbf{69.90} & \textbf{65.65} \\
    \bottomrule
    \end{tabular}
    }
    \vspace{-0.5em}
    \caption{Top-1 classification accuracy in $\%$ of six model variants. ``P-DNN'' denotes the plain DNN. Bold font indicates the best performance, and blue font denotes the second best. Experiments are repeated three times and the average results are reported.}
    \label{tab:vam-over-vanilla-kd}
    \vspace{-1em}
\end{table*}

To understand the effects of vanilla KD, KDExplainer is trained with and without soft targets as the optimization objective. Furthermore, as label smoothing~\cite{Szegedy2016RethinkingTI}  is
recently also viewed as a type of KD~\cite{Yuan_2020_CVPR}, we also train the KDExplainer with label smoothing to understand its effects. In Figure~\ref{fig:experiment-tree}, we visualize the derived neural trees by keeping only the maximum attention weight in every attention module as Eqn.~\ref{eq:keep-maximum}. It can be seen that KD significantly encourages sparse connectivity between blocks, as its retained blocks are much fewer than normal training and label smoothing. It can be also verified by the attention distribution shown in the histograms in internal nodes, where KD produces sharper~(i.e., lower entropy) attention distribution than normal training and label smoothing. Furthermore, KD produces more human-interpretable branching architecture than normal training and label smoothing. For example, in the derived trees, man-made vehicle categories are attached to the same branch, while animal categories are attached to other branches. This is a reasonable organization as similar categories or tasks share similar decision patterns, and thus they can share the same network branch. Less related categories or tasks, on the other hand, depend on different patterns to make their decision, such that they should be solved separately. The results of normal training and label smoothing somewhat violate this principle, thus their derived trees are much larger than that of KD.
\begin{figure}[t]
     \centering
     \subfigure[CIFAR-10]{
    \includegraphics[scale=0.595]{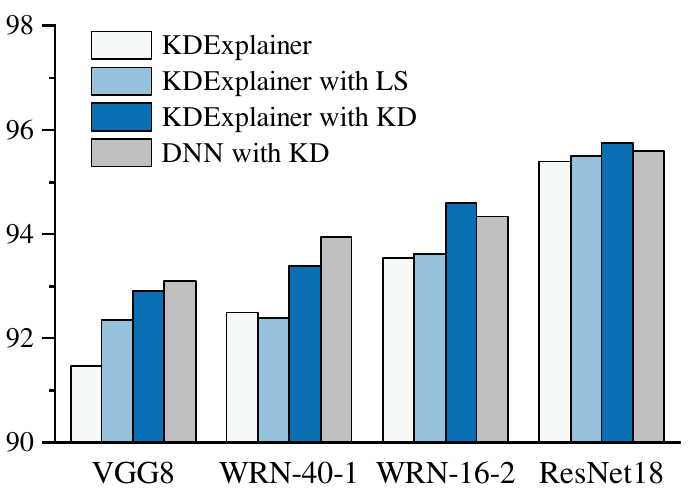}}
     \subfigure[CIFAR-100]{
    \includegraphics[scale=0.595]{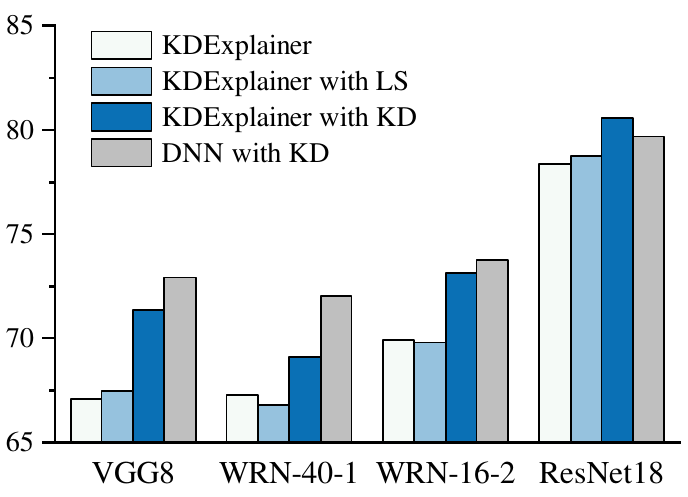}}
    \vspace{-1em}
    \caption{Accuracy~(\%) of KDExplainers trained with different objectives in different architectures. ``DNN with KD'' and ``LS'' denote the conventional DNN trained with KD and label smoothing.}
    \vspace{-1em}
    \label{fig:kde-10-100}
\end{figure}


In Figure~\ref{fig:kde-10-100}, we provide the accuracies of all the KDExplainers. Under all experimental settings, KD significantly outperforms label smoothing and vanilla training. Label smoothing marginally improves the performance in most cases compared to vanilla training, but it sometimes leads to performance degradation. For example, it causes performance drops by $0.51\%$ and $0.14\%$ of WRN-40-1 and WRN-16-2 on CIFAR-100.  These results prove that by properly organizing categories into different branches, KD helps modulate the knowledge conflicts between different tasks and thus achieves higher accuracy. Label smoothing, on the other hand, provides some regularization on the model learning, but it seems to have play no role  in addressing  conflicts between decision patterns of different categories. In the supplementary material, we provide more results and analyses to explain the differences between KD and label smoothing.

\subsection{Improving KD with VAM}
\label{subsec:experiment-VAM}
\subsubsection{Experimental settings}
Experiments are conducted on  CIFAR-10, CIFAR-100, and Tiny-ImageNet~\cite{le2015tiny}. We incorporate VAM into six widely-used DNNs as student models, including ResNet18, resnet20\footnote{Fowlloing~\cite{tian2019crd}, ResNet and resnet represent cifar- and ImageNet-style networks, respectively.}, VGG8, WRN-16-2, WRN-40-1, and ShuffleNetV1~\cite{zhang2018shufflenet}. For all involved student models, we adopt ResNet50 as their teacher model to provide the soft logits. During the training phase of the student model, the learning rate is initially $0.05$ (attention module $0.01$), gets decayed at epoch $150$, $180$, $210$ by a factor of $0.1$, and ceases at $240$. The temperature hyper-parameter is set to $1$ for all attention modules and $4$ for KD loss. The trading-off factor $\alpha$ is set to $0.9$. The number of channels in each virtual block is $8$ for VGG8, WRN-16-2, and WRN-40-1, $4$ for resnet20, $16$ for ResNet18, and $10$ for ShuffleNetV1. For more details, please refer to the supplementary material.



\subsubsection{Experimental Results}
\textbf{Improving vanilla KD.}~~As VAM is motivated by vanilla KD, we first validate its effectiveness upon vanilla KD. To give a more comprehensive view of VAM, we make comparisons between six model variants. Note that we simply use VAM to denote the DNN incorporated with the proposed virtual attention module. (M1) Plain+$\mathcal{L}_{\text{CE}}$: plain DNN trained with only $\mathcal{L}_{\text{CE}}$; (M2) VAM+$\mathcal{L}_{\text{CE}}$: VAM trained with $\mathcal{L}_{\text{CE}}$; (M3) VAM+$\mathcal{L}_{\text{CE}}$+$\mathcal{H}$: VAM trained with $\mathcal{L}_{\text{CE}}$ and $\mathcal{H}$ (Eqn.~\ref{eq:entropy-minimize}); (M4) Plain+$\mathcal{L}_{\text{CE}}$+$\mathcal{D}_{\text{KL}}$: plain DNN trained with $\mathcal{L}_{\text{CE}}$ and $\mathcal{D}_{\text{KL}}$; (M5) VAM+$\mathcal{L}_{\text{CE}}$+$\mathcal{D}_{\text{KL}}$: VAM trained with $\mathcal{L}_{\text{CE}}$ and $\mathcal{D}_{\text{KL}}$; (M6) VAM+$\mathcal{L}_{\text{CE}}$+$\mathcal{D}_{\text{KL}}$+$\mathcal{H}$: VAM trained with  $\mathcal{L}_{\text{CE}}$, $\mathcal{D}_{\text{KL}}$ and $\mathcal{H}$. Experimental results are shown in Table~\ref{tab:vam-over-vanilla-kd}. In general, DNN incorporated with  VAM yields consistently superior performance to the plain DNN without VAM. For example, M5 consistently outperforms M4 in almost all our experiments, which validates its effectiveness under KD. Furthermore, when optimized with low entropy constraint $\mathcal{H}$, VAM produces better performance (M3$>$M2, M6$>$M5) under almost all settings. As $\mathcal{H}$ is motivated by the results from KDExplainer, it indicates that the proposed KDExplainer indeed provides us with general and valuable insights into KD.
\noindent\textbf{Improving state-of-the-art KD.}
\begin{table}[t]
\centering
\resizebox{0.48\textwidth}{!}{   
\begin{tabular}{cc|cc|cc}
  \toprule
   & & \multicolumn{2}{c|}{\textbf{CIFAR-10}} & \multicolumn{2}{c}{\textbf{CIFAR-100}}  \\
   \textbf{Method} &\textbf{Model} & resnet20 & WRN-16-2 & resnet20 & WRN-16-2  \\  \midrule
  \multirow{2}{*}{FitNet} & \small{P-DNN} & 92.49 & 93.99 & 69.09 & 72.98  \\
   & \small{VAM} & \textbf{92.95} & \textbf{94.08} & \textbf{70.11} &  \textbf{73.79}
  \\ \midrule
  \multirow{2}{*}{FT} & \small{P-DNN} & 92.39 & 93.74 & 69.47 & 72.78 \\ 
   & \small{VAM} & \textbf{93.01} & \textbf{93.78} & \textbf{69.67} & \textbf{73.43}
  \\ \midrule
  \multirow{2}{*}{SP} & \small{P-DNN} & 93.01 & 94.55 & 69.17 & 74.40
  \\
   & \small{VAM} & \textbf{93.51} & \textbf{94.61} & \textbf{70.35} & \textbf{74.91} \\ \midrule
   \multirow{2}{*}{CRD} & \small{P-DNN} & 91.83 & 93.36 & 70.64 & 74.97 \\
   & \small{VAM} & \textbf{92.16} & \textbf{93.53} & \textbf{70.85} & \textbf{75.15} \\ 
 \bottomrule
\end{tabular}
}
\vspace{-0.5em}
\caption{Results of combing VAM with other KD methods. ``P-DNN'' denotes the plain DNN.}
\vspace{-1em}
\label{tab:vam-sota}
\end{table}
~~Although VAM is motivated by vanilla KD, it is also straightforward to be combined with other KD methods. Here we evaluate VAM combined with other KD methods, including FitNets~\cite{romero2014fitnets}, FT~\cite{kim2018paraphrasing}, SP~\cite{tung2019similarity}, and CRD~\cite{tian2019contrastive}. Results are listed in Table~\ref{tab:vam-sota}. It can be seen that combined with other KD methods, VAM still yields  performances consistently superior to the plain DNN though it is motivated by only vanilla KD. For resnet20, VAM achieves $0.20\%\sim1.18\%$ performance boosts on CIFAR-100 in four KD methods. For WRN-16-2, VAM improves $0.18\%\sim0.81\%$ on CIFAR-100. Since VAM brings negligible additional overhead, it is an economical way to further improve the performance of existing KD methods. Please refer to the supplementary material for more experimental results, including how the block size and the hyper-parameter $\gamma$ affect the final performance.

\section{Conclusion and Future Work}
In this paper, we propose KDExplainer to shed light on 
the working mechanism underlying soft targets
during KD. We find that KD implicitly modulates 
the knowledge conflicts between different subtasks, 
and effectively brings about more benefits as compared to label smoothing. 
Based on these observations, we propose a portable module,  VAM,
to further improve the results of 
vanilla KD alongside other state-of-the-art ones. 
Extensive experimental results exhibit that
the proposed VAM significantly enhances the KD performance
at a negligible additional cost. In our future work, we will extend the proposed VAM to other tasks and systematically evaluate its effectiveness beyond the scope of KD.

\paragraph{Acknowledgment}
This work is funded by the National Key R\&D Program of China (Grant No: 2018AAA0101503) and the Science and technology project of SGCC (State Grid Corporation of China): fundamental theory of human-in-the-loop hybrid-augmented intelligence for power grid dispatch and control.

{\small
\bibliographystyle{named}
\bibliography{ijcai21}
}

\end{document}